\documentclass[conference]{IEEEtran}
\IEEEoverridecommandlockouts
\usepackage{cite}
\usepackage{amsmath,amssymb,amsfonts}
\usepackage{algorithmic}
\usepackage{graphicx}
\usepackage{textcomp}
\usepackage{multirow}
\usepackage{tabularx}
\usepackage{makecell}
\usepackage{hyperref}
\usepackage{xcolor}
\usepackage{caption}
\usepackage{subcaption}
\usepackage{tikz,pgfplots}
\usetikzlibrary{positioning}
\usetikzlibrary{arrows}
\usetikzlibrary{patterns}
\usetikzlibrary{shapes.geometric}

\def\BibTeX{{\rm B\kern-.05em{\sc i\kern-.025em b}\kern-.08em
    T\kern-.1667em\lower.7ex\hbox{E}\kern-.125emX}}
\begin{document}

\title{Rethinking RAFT for Efficient Optical Flow \\
}

\author{
  \IEEEauthorblockN{Navid Eslami, Farnoosh Arefi, Amir M. Mansourian, Shohreh Kasaei}
  \IEEEauthorblockA{
    Department of Computer Engineering\\
   Sharif University of Technology\\
    Tehran, Iran\\
    Email: \{navid.eslami, far.arefi, amir.mansurian, kasaei\}@sharif.edu
  }

}

\maketitle

\maketitle

\begin{abstract}
Despite significant progress in deep learning-based optical flow methods, accurately estimating large displacements and repetitive patterns remains a challenge. The limitations of local features and similarity search patterns used in these algorithms contribute to this issue. Additionally, some existing methods suffer from slow runtime and excessive graphic memory consumption. To address these problems, this paper proposes a novel approach based on the RAFT framework. The proposed Attention-based Feature Localization (AFL) approach incorporates the attention mechanism to handle global feature extraction and address repetitive patterns. It introduces an operator for matching pixels with corresponding counterparts in the second frame and assigning accurate flow values. Furthermore, an Amorphous Lookup Operator (ALO) is proposed to enhance convergence speed and improve RAFT's ability to handle large displacements by reducing data redundancy in its search operator and expanding the search space for similarity extraction. The proposed method, Efficient RAFT (Ef-RAFT), achieves significant improvements of 10\% on the Sintel dataset and 5\% on the KITTI dataset over RAFT. Remarkably, these enhancements are attained with a modest 33\% reduction in speed and a mere 13\% increase in memory usage. The code is available at: \textnormal{\href{https://github.com/n3slami/Ef-RAFT}{https://github.com/n3slami/Ef-RAFT}}


\end{abstract}

\begin{IEEEkeywords}
Optical Flow, Large Displacement, Repetitive Patterns, Attention Mechanism, Deep Neural Networks
\end{IEEEkeywords}

\section{Introduction}

Optical Flow is a fundamental challenge in computer vision, focusing on determining the displacement vector for each pixel between two consecutive frames. This technique holds immense significance across downstream tasks, like visual tracking\cite{vihlman2020optical}, video segmentation\cite{yang2021self}, and robot navigation\cite{de2021enhancing}. Traditionally, the problem has been tackled using different classical computer vision methods such as correlation-based\cite{singh1991optic}, block matching\cite{beauchemin1995computation}, and energy minimization-based\cite{horn1981determining} techniques. However, these approaches have proven to be computationally expensive, making them impractical for real-time applications.

In recent years, deep learning has emerged as a promising alternative to conventional approaches. Deep learning techniques have the advantage of bypassing the need to formulate optimization problems and instead train networks to directly predict the optical flow. These methods\cite{dosovitskiy2015flownet, ranjan2017optical, sun2018pwc} have demonstrated comparable performance to the top traditional methods while significantly reducing the inference time, making them faster and more efficient. In general, neural network models take a pair of consecutive images captured by a frame-based camera as input and generate predictions for the optical flow that effectively warps pixels from one image to the other. 

RAFT\cite{teed2020raft}, which stands for Recurrent All-Pairs Field Transforms, is considered one of the most successful learning-based methods for optical flow estimation. It has gained significant popularity as a simple yet robust baseline approach in the field. While this method exhibits efficient performance when evaluated on benchmark datasets, it can still encounter challenges under specific conditions. For instance, when dealing with significant displacements or untextured/repetitive patterns, there is a possibility of encountering large errors in the estimated optical flow. To enhance the performance of optical flow estimation, advanced techniques have been developed specifically for enhancing the RAFT method. These techniques encompass attention-based operations\cite{jiang2021learning, zhao2022global}, graph models\cite{luo2022learning}, and latent cost-volume augmentation\cite{huang2022flowformer}. However, it is important to note that these methods typically require additional computational resources and introduce significant inference time, which limits their practical application in real-world scenarios. 

To tackle the challenges posed by large displacements and repetitive patterns in the RAFT method, this paper introduces two novel mechanisms: 1) the Amorphous Lookup Operator (ALO), and 2) the Attention-based Feature Localizer (AFL). The former replaces the original lookup operator of RAFT, allowing the network to vary the distribution of the correlation queries based on the input frames, resulting in the ability to recognize larger displacements. The latter transforms the resultant features of the feature encoders in such a way that the network can differentiate and match the pixels in a poorly textured region, reducing the ambiguities that stem from these regions and thus, mitigating the resultant errors in estimation.

In summary, the main contributions of this work are as
follows:

\begin{itemize}
  \item Proposing the Amorphous Lookup Operator (ALO) as a novel method for tackling large displacement problem of the optical flow estimation methods. 

  \item Proposing the Attention-based Feature Localizer (AFL) as a novel method for tackling repetitive patterns problem of the optical flow estimation methods.

  \item Validating the effectiveness of the proposed methods by conducting extensive experiments on the Sintel and KITTI datasets.

\end{itemize}

\section{RELATED WORK}
Traditionally, optical flow was commonly approached using energy-based methods\cite{horn1981determining, zach2007duality, black1993framework} that employed a variational approach by defining a data term and a regularization term. To address the challenge of large displacements, improved estimation methods such as pyramid approaches\cite{brox2004high} were introduced. Additionally, feature-based methods emerged as a solution to mitigate the large displacement problem\cite{bailer2015flow, menze2015discrete, weinzaepfel2013deepflow, brox2010large, revaud2015epicflow}. These methods defined a matching term and utilized dynamic programming and interpolation techniques to enhance accuracy and robustness.

Significant progress has been made in the field of optical flow estimation, thanks to recent advancements in deep learning. FlowNet\cite{dosovitskiy2015flownet} introduced the first end-to-end network capable of tackling this task. Subsequently, a series of methods based on neural networks\cite{ilg2017flownet, hui2018liteflownet, sun2018pwc, ranjan2017optical, zhao2020maskflownet, hur2019iterative, yang2019volumetric, deng2023explicit, zhang2021separable} have been developed, aiming to enhance performance through coarse-to-fine or iterative approaches. These methods have contributed to a notable improvement in optical flow estimation. Typically, these methods employ a correlation matrix to capture the similarities between pixels in consecutive frames. This correlation matrix takes the form of a 4D volume, which can be quite large depending on the size of the frames being processed. Due to the limited receptive field of Convolutional Neural Networks (CNNs), transformer-based approaches have recently emerged\cite{huang2022flowformer, lu2023transflow, shi2023flowformer++} as a solution. These transformer architectures have shown superior performance compared to CNNs. However, it is worth noting that transformer-based approaches often require a substantial amount of data and annotations to achieve their optimal performance.

RAFT\cite{teed2020raft}, drawing inspiration from variational methods, incorporates recurrent gate units and correlation matrices at multiple scales. It stimulates an iterative optimization problem that improves the flow estimations through iterative refinement. Following the utilization of large correlation matrices in RAFT, numerous efforts have been made to enhance the performance of RAFT in various aspects\cite{jiang2021learning, zhao2022global, luo2022learning, xu2021high, sui2022craft}. Many of the mentioned works focus on leveraging the attention mechanism to improve RAFT. In particular,\cite{xu2021high} introduces an attention-based approach to estimate the similarity between pairs of pixels. By incorporating vertical and horizontal correlations, it enhances the estimation process. Notably,\cite{huang2022flowformer} replaces correlation matrices with correlation memory, which consists of tokens describing pixels. Correlation values are computed using attention within this memory structure. Additionally,\cite{sui2022craft} employs the attention mechanism to address the impact of noise on correlation matrices, resulting in the extraction of more globally coherent and semantically stable features. Occlusion poses a significant challenge in optical flow estimation, and\cite{jiang2021learning} addresses this issue by employing attention to generalize information to occluded areas. By utilizing attention, it aims to provide a more comprehensive understanding of the scene, even in the presence of occlusions. Repetitive patterns, on the other hand, introduce difficulties in determining corresponding pixels due to their similar feature vectors.\cite{zhao2022global} utilizes attention to identify pixels that are highly likely to be correct correspondences, which proves beneficial in handling scenarios with large displacements. By leveraging attention, it enhances the ability to cope with repetitive patterns and improve the accuracy of optical flow estimation.

In this paper, building upon the RAFT framework, two novel methods are introduced to specifically tackle the challenges posed by large displacements and repetitive patterns. These methods, referred to as Attention-based Feature Localizer and  Amorphous Lookup Operator significantly enhance the performance of RAFT. Notably, these proposed techniques achieve notable improvements while imposing a minimal burden on memory consumption and runtime.

\section{PROPOSED METHOD}
In this section, first the optical flow baseline, RAFT\cite{teed2020raft}, is reviewed as this work is based on the RAFT architecture. Subsequently, a detailed explanation of the two proposed methods for addressing large displacements and repetitive patterns are provided.

\begin{figure*}[!ht]
\centerline{\includegraphics[scale=.5]{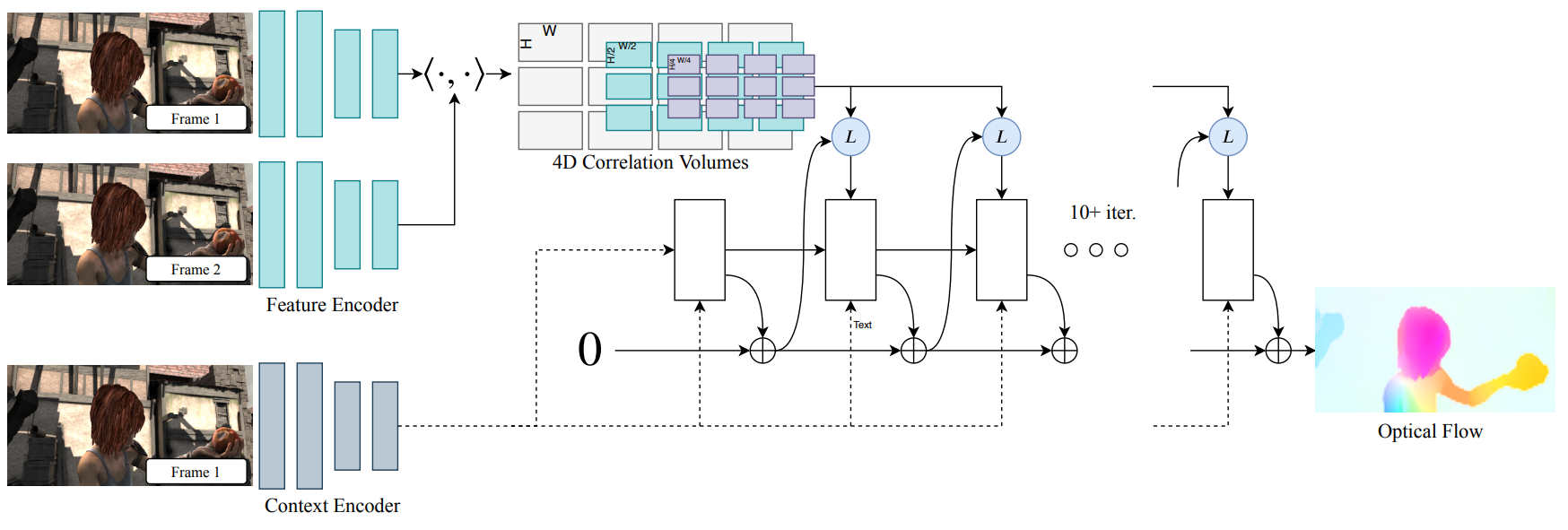}}
\caption{Diagram of RAFT\cite{teed2020raft} encompasses three main components. 1) Feature encoders are employed to extract per-pixel features from the input frames. 2) A correlation layer constructs a correlation volume with dimensions $W \times H \times W \times H$ by computing the inner product of feature vectors for all pairs. 3) An update operator recurrently enhances the optical flow estimation by leveraging the current estimate to retrieve values from the set of correlation volumes.}
\label{fig:1_RAFT_Diagram}
\end{figure*}

\subsection{Optical Flow Estimation Baseline}
Given a pair of consecutive frames, $I_1$ and $I_2$, a dense displacement field $(a_1, a_2)$ is estimated. This field maps each pixel $(u, v)$ in $I_1$ to its corresponding coordinates $(u', v') = (u + a_1(u), v + a_2(v))$ in $I_2$. As illustrated in Fig\ref{fig:1_RAFT_Diagram}, the RAFT model consists of some main components, each of which is briefly explained below. For more information about the RAFT baseline, refer to\cite{teed2020raft}.

\subsubsection{Encoders}
Two encoders with residual blocks are used in RAFT to extract features and reduce noise. As depicted in Fig \ref{fig:1_RAFT_Diagram}, the feature encoder, $g_\theta$, is employed to extract features from both frames. On the other hand, the context encoder, $h_\theta$, is applied to $I_1$ to extract features for optimization and estimate the flow.

\subsubsection{Correlation Volume}
This volume aims to represent the similarities between the points in the pair of frames. Let's suppose that $F_1=g_\theta(I_1)$ and $F_2=g_\theta(I_2)$ are the extracted features from the feature encoder for $I_1$ and $I_2$, respectively. The correlation volume, $C$, can be calculated as follows\cite{teed2020raft}:

\begin{equation} \label{Correlation_Volume} 
    \centering
C^l_{i,j,x,y} =  \begin{cases} \langle F_1(i,j), F_2(x,y)\rangle & l = 0 \\
                     Average Pooling_{x, y}(C^{l-1}_{i,j,x,y}) &  l\ge 1.
       \end{cases}
\end{equation}

Where $C^l$ denotes the correlation at the $l$-th level. In fact, the first level ($l=0$) is created by calculating the dot product of each pair of points in $F_1$ and $F_2$. The next three levels are formed by applying average pooling on the last two dimensions of the correlation volume in the former layer.

\subsubsection{Lookup Operator}
Given a current estimate of optical flow ($a_1$, $a_2$), each pixel $p = (u, v)$ in $I_1$ is mapped to its estimated correspondence in $I_2$: $q = (u + a_1(u), v + a_2(v))$. Following that, a local grid is defined around $q$:

\begin{equation} \label{Grid} 
    \centering
    \mathcal{N}(q)_r = \{q+h | h \in \mathbb{Z}^2, ||h||_1 \le r\} 
\end{equation}  

where $r$ is the radius of the grid. This operator extracts similarity values from the correlation volume as a vector for a pair of points, p and q.

\subsubsection{Recurrent Gate Units}
This component simulates the optimization algorithm, which estimates a sequence of flow estimates $\{a_1, a_2, ..., a_N\}$ starting from an initial flow $a_0=0$. It takes the current flow, correlations, and a hidden state as input, and iteratively updates the flow and hidden state in output.

Finally, the loss function of the network is defined as follows:
\begin{equation} \label{Loss Function} 
    \centering
    \ell = \sum_{i=1}^{N} \gamma^{N-i}||a_{gt} - a_i||_1 
\end{equation}  

Where $a_{gt}$ represents the ground truth flows, $a_i$ denotes the calculated flows in the $i$-th step of the algorithm and $\gamma$ is a hyperparameter.

\subsection{Amorphous Lookup Operator}
RAFT's vanilla lookup operator has three major short-comings: 1) It's structure imposes a hard limit of 256 pixels on the distance that the network can gather information from in each iteration. 2) In practice it often fails to find meaningful correspondences up to this range. 3) Since the lookup operator is used on all levels of the correlation volume, the data extracted is pooled from the same regions, causing unneeded redundancy. The idea of the Amorphous Lookup Operator (ALO) is to allow the network to change the lookup operator so that, based on the input frames, it can choose to query farther away similarities and reduce the redundancy of the extracted similarities.

\begin{figure}
\centering
\begin{subfigure}[b]{0.2\columnwidth}
\centering
\hspace{-2\columnwidth} 
\begin{tikzpicture}
\filldraw[blue!10] (-1.2, -1.2) rectangle (1.2, 1.2);
\foreach \x in {-4, ..., 4}
{
	\draw[blue] (0.3 * \x, -1.2) -- +(0, 2.4);
	\draw[blue] (-1.2, 0.3 * \x) -- +(2.4, 0);
	\foreach \y in {-4, ..., 4}
	{
		\filldraw[blue] (0.3 * \x, 0.3 * \y) circle [radius=1pt];
	}
}
\end{tikzpicture}
\end{subfigure}
\begin{subfigure}[b]{0.2\columnwidth}
\centering
\begin{tikzpicture}
\def\dx{1};
\def\dy{0.75};
\filldraw[blue] (0, 0) circle [radius=1pt];
\filldraw[blue!10] (\dx, -\dy) rectangle +(1.05, -1.5);
\filldraw[blue!10] (\dx, \dy) rectangle +(1.05, 1.5);
\filldraw[blue!10] (-\dx, -\dy) rectangle +(-1.05, -1.5);
\filldraw[blue!10] (-\dx, \dy) rectangle +(-1.05, 1.5);

\draw[blue] (\dx, 0) -- +(1.05, 0);
\draw[blue] (-\dx, 0) -- +(-1.05, 0);
\draw[blue] (0, \dy) -- +(0, 1.5);
\draw[blue] (0, -\dy) -- +(0, -1.5);

\foreach \x in {0, ..., 3}
{
	\filldraw[blue] (\dx + 0.35 * \x, 0) circle [radius=1pt];	
	\filldraw[blue] (-\dx + -0.35 * \x, 0) circle [radius=1pt];	
	\filldraw[blue] (0, \dy + 0.5 * \x) circle [radius=1pt];
	\filldraw[blue] (0, -\dy - 0.5 * \x) circle [radius=1pt];	
	
	\draw[blue] (\dx + 0.35 * \x, \dy) -- +(0, 1.5);
	\draw[blue] (\dx, \dy + 0.5 * \x) -- +(1.05, 0);
	
	\draw[blue] (-\dx - 0.35 * \x, \dy) -- +(-0, 1.5);
	\draw[blue] (-\dx, \dy + 0.5 * \x) -- +(-1.05, 0);
	
	\draw[blue] (\dx + 0.35 * \x, -\dy) -- +(0, -1.5);
	\draw[blue] (\dx, -\dy + -0.5 * \x) -- +(1.05, 0);
	
	\draw[blue] (-\dx - 0.35 * \x, -\dy) -- +(-0, -1.5);
	\draw[blue] (-\dx, -\dy + -0.5 * \x) -- +(-1.05, 0);
	\foreach \y in {0, ..., 3}
	{
		\filldraw[blue] (\dx + 0.35 * \x, \dy + 0.5 * \y) circle [radius=1pt];
		\filldraw[blue] (-\dx - 0.35 * \x, \dy + 0.5 * \y) circle [radius=1pt];
		\filldraw[blue] (\dx + 0.35 * \x, -\dy + -0.5 * \y) circle [radius=1pt];
		\filldraw[blue] (-\dx - 0.35 * \x, -\dy + -0.5 * \y) circle [radius=1pt];
	}
}

\draw[-, red, thick, decorate, decoration={brace, amplitude=3, aspect=0.5}] (0, \dy + 1.5) -- +(\dx, 0) node[pos=0.5, above=0pt, black]{$+d_x$};
\draw[-, red, thick, decorate, decoration={brace, mirror, amplitude=3, aspect=0.5}] (0, \dy + 1.5) -- +(-\dx, 0) node[pos=0.5, above=0pt, black]{$-d_x$};

\draw[-, red, thick, decorate, decoration={brace, mirror, amplitude=3, aspect=0.5}] (\dx + 1.05, 0) -- +(0, \dy) node[pos=0.5, rotate=-90, above=0pt, black]{$+d_y$};
\draw[-, red, thick, decorate, decoration={brace, amplitude=3, aspect=0.5}] (\dx + 1.05, 0) -- +(0, -\dy) node[pos=0.5, rotate=-90, above=0pt, black]{$-d_y$};
\end{tikzpicture}
\end{subfigure}

\caption{Structure of the grid used in the original lookup operator (left), compared to the transformed lookup operator used in the ALO (right).}
\label{fig:asl_operator}
\end{figure}
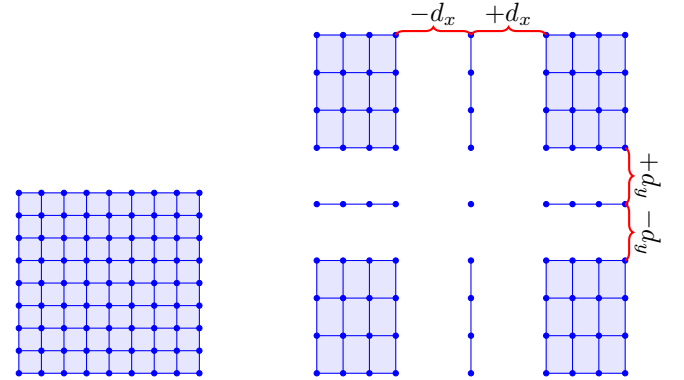

The ALO changes the original grids of the lookup operator by scaling and translating parts of it. That is, it first calculates four scalar parameters $s_x, s_y \in [1, 3]$, $d_x, d_y \in [0, 2]$ based on the input frames. It then transforms each point $p_{old} = (i, j)$ in the old lookup operator to 
\begin{equation}
p_{new} = (s_x i + \textbf{sign}(i) d_x, s_y j + \textbf{sign}(j) d_y), 
\end{equation}
where $\textbf{sign}(\cdot)$ is a function returning $+1$ for positive inputs, $-1$ for negative inputs, and $0$ for an input of zero. This transformation breaks the grids into four sub-grids, which can be positioned and spaced arbitrarily, while maintaining symmetry, as shown in Fig.~\ref{fig:asl_operator}. The parameters $s_x$, $s_y$, $d_x$ and $d_y$ are calculated separately for each level of the correlation volume, which enable the network to adaptively pool similarity data from farther regions, while avoiding the extraction of too much redundant information.

\begin{figure}
\resizebox{\columnwidth}{!}{%
\begin{tikzpicture}

\node[inner sep=0pt] (frame_1_context) at (-2, -7) {\includegraphics[width=.20\textwidth]{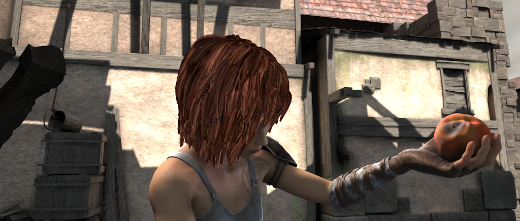}};
\node[draw=black, fill=white, thick, rounded corners=.15cm, inner sep=3pt, left=-0.5cm of frame_1_context]  {Frame 1};

\node (fcon_1_l1) [above right=0.05cm of frame_1_context] {};
\filldraw[blue!70, opacity=0.1] (fcon_1_l1) rectangle +(0.25, -1.7);
\draw[blue!70] (fcon_1_l1) rectangle +(0.25, -1.7);
\node (fcon_1_l2) [right=0.1cm of fcon_1_l1] {};
\filldraw[blue!70, opacity=0.1] (fcon_1_l2) rectangle +(0.25, -1.7);
\draw[blue!70] (fcon_1_l2) rectangle +(0.25, -1.7);
\node (fcon_1_l3) [below right=-0.1cm and 0.1cm of fcon_1_l2] {};
\filldraw[blue!70, opacity=0.1] (fcon_1_l3) rectangle +(0.25, -1.3);
\draw[blue!70] (fcon_1_l3) rectangle +(0.25, -1.3);
\node (fcon_1_l4) [right=0.1cm of fcon_1_l3] {};
\filldraw[blue!70, opacity=0.1] (fcon_1_l4) rectangle +(0.25, -1.3);
\draw[blue!70] (fcon_1_l4) rectangle +(0.25, -1.3);

\node[below right=0.1cm and -0.25cm of frame_1_context]  {Context Encoder};

\node[gray!10, fill, minimum width=1cm, minimum height=0.6cm] (cv0_00) at (1, 4) {};
\node[inner sep=0pt, gray, draw, thick, minimum width=1cm, minimum height=0.6cm] at (cv0_00) {\scalebox{.65}{$H\times W$} \normalsize};
\node[gray!10, right=0.1cm of cv0_00, fill, minimum width=1cm, minimum height=0.6cm] (cv0_01) {};
\node[gray, draw, thick, minimum width=1cm, minimum height=0.6cm] at (cv0_01) {};
\node[gray!10, right=0.1cm of cv0_01, fill, minimum width=1cm, minimum height=0.6cm] (cv0_02) {};
\node[gray, draw, thick, minimum width=1cm, minimum height=0.6cm] at (cv0_02) {};
\node[gray!10, right=0.1cm of cv0_02,, fill, minimum width=1cm, minimum height=0.6cm] (cv0_03) {};
\node[gray, draw, thick, minimum width=1cm, minimum height=0.6cm] at (cv0_03) {};

\node[gray!10, below=0.1cm of cv0_00, fill, minimum width=1cm, minimum height=0.6cm] (cv0_10) {};
\node[gray, draw, thick, minimum width=1cm, minimum height=0.6cm] at (cv0_10) {};
\node[gray!10, right=0.1cm of cv0_10, fill, minimum width=1cm, minimum height=0.6cm] (cv0_11) {};
\node[gray, draw, thick, minimum width=1cm, minimum height=0.6cm] at (cv0_11) {};
\node[gray!10, right=0.1cm of cv0_11, fill, minimum width=1cm, minimum height=0.6cm] (cv0_12) {};
\node[gray, draw, thick, minimum width=1cm, minimum height=0.6cm] at (cv0_12) {};
\node[gray!10, right=0.1cm of cv0_12, fill, minimum width=1cm, minimum height=0.6cm] (cv0_13) {};
\node[gray, draw, thick, minimum width=1cm, minimum height=0.6cm] at (cv0_13) {};

\node[gray!10, below=0.1cm of cv0_10, fill, minimum width=1cm, minimum height=0.6cm] (cv0_20) {};
\node[gray, draw, thick, minimum width=1cm, minimum height=0.6cm] at (cv0_20) {};
\node[gray!10, right=0.1cm of cv0_20, fill, minimum width=1cm, minimum height=0.6cm] (cv0_21) {};
\node[gray, draw, thick, minimum width=1cm, minimum height=0.6cm] at (cv0_21) {};
\node[gray!10, right=0.1cm of cv0_21, fill, minimum width=1cm, minimum height=0.6cm] (cv0_22) {};
\node[gray, draw, thick, minimum width=1cm, minimum height=0.6cm] at (cv0_22) {};
\node[gray!10, right=0.1cm of cv0_22, fill, minimum width=1cm, minimum height=0.6cm] (cv0_23) {};
\node[gray, draw, thick, minimum width=1cm, minimum height=0.6cm] at (cv0_23) {};

\node[cyan!10, below right=-0.1cm and 0.15cm of cv0_00, fill, minimum width=0.8cm, minimum height=0.48cm] (cv1_00) {};
\node[inner sep=0pt, cyan, draw, thick, minimum width=0.8cm, minimum height=0.48cm] at (cv1_00) {\scalebox{.53}{$\frac{H}{2}\times \frac{W}{2}$} \normalsize};
\node[cyan!10, right=0.1cm of cv1_00, fill, minimum width=0.8cm, minimum height=0.48cm] (cv1_01) {};
\node[cyan, draw, thick, minimum width=0.8cm, minimum height=0.48cm] at (cv1_01) {};
\node[cyan!10, right=0.1cm of cv1_01, fill, minimum width=0.8cm, minimum height=0.48cm] (cv1_02) {};
\node[cyan, draw, thick, minimum width=0.8cm, minimum height=0.48cm] at (cv1_02) {};
\node[cyan!10, right=0.1cm of cv1_02, fill, minimum width=0.8cm, minimum height=0.48cm] (cv1_03) {};
\node[cyan, draw, thick, minimum width=0.8cm, minimum height=0.48cm] at (cv1_03) {};

\node[cyan!10, below=0.1cm of cv1_00, fill, minimum width=0.8cm, minimum height=0.48cm] (cv1_10) {};
\node[cyan, draw, thick, minimum width=0.8cm, minimum height=0.48cm] at (cv1_10) {};
\node[cyan!10, right=0.1cm of cv1_10, fill, minimum width=0.8cm, minimum height=0.48cm] (cv1_11) {};
\node[cyan, draw, thick, minimum width=0.8cm, minimum height=0.48cm] at (cv1_11) {};
\node[cyan!10, right=0.1cm of cv1_11, fill, minimum width=0.8cm, minimum height=0.48cm] (cv1_12) {};
\node[cyan, draw, thick, minimum width=0.8cm, minimum height=0.48cm] at (cv1_12) {};
\node[cyan!10, right=0.1cm of cv1_12, fill, minimum width=0.8cm, minimum height=0.48cm] (cv1_13) {};
\node[cyan, draw, thick, minimum width=0.8cm, minimum height=0.48cm] at (cv1_13) {};

\node[cyan!10, below=0.1cm of cv1_10, fill, minimum width=0.8cm, minimum height=0.48cm] (cv1_20) {};
\node[cyan, draw, thick, minimum width=0.8cm, minimum height=0.48cm] at (cv1_20) {};
\node[cyan!10, right=0.1cm of cv1_20, fill, minimum width=0.8cm, minimum height=0.48cm] (cv1_21) {};
\node[cyan, draw, thick, minimum width=0.8cm, minimum height=0.48cm] at (cv1_21) {};
\node[cyan!10, right=0.1cm of cv1_21, fill, minimum width=0.8cm, minimum height=0.48cm] (cv1_22) {};
\node[cyan, draw, thick, minimum width=0.8cm, minimum height=0.48cm] at (cv1_22) {};
\node[cyan!10, right=0.1cm of cv1_22, fill, minimum width=0.8cm, minimum height=0.48cm] (cv1_23) {};
\node[cyan, draw, thick, minimum width=0.8cm, minimum height=0.48cm] at (cv1_23) {};

\node[blue!10, below right=-0.15cm and 0.06cm of cv1_00, fill, minimum width=0.64cm, minimum height=0.45cm] (cv2_00) {};
\node[inner sep=0pt, blue, draw, thick, minimum width=0.64cm, minimum height=0.45cm] at (cv2_00) {\scalebox{.4}{$\frac{H}{4}\times \frac{W}{4}$} \normalsize};
\node[blue!10, right=0.05cm of cv2_00, fill, minimum width=0.64cm, minimum height=0.45cm] (cv2_01) {};
\node[blue, draw, thick, minimum width=0.64cm, minimum height=0.45cm] at (cv2_01) {};
\node[blue!10, right=0.05cm of cv2_01, fill, minimum width=0.64cm, minimum height=0.45cm] (cv2_02) {};
\node[blue, draw, thick, minimum width=0.64cm, minimum height=0.45cm] at (cv2_02) {};
\node[blue!10, right=0.05cm of cv2_02, fill, minimum width=0.64cm, minimum height=0.45cm] (cv2_03) {};
\node[blue, draw, thick, minimum width=0.64cm, minimum height=0.45cm] at (cv2_03) {};

\node[blue!10, below=0.05cm of cv2_00, fill, minimum width=0.64cm, minimum height=0.45cm] (cv2_10) {};
\node[blue, draw, thick, minimum width=0.64cm, minimum height=0.45cm] at (cv2_10) {};
\node[blue!10, right=0.05cm of cv2_10, fill, minimum width=0.64cm, minimum height=0.45cm] (cv2_11) {};
\node[blue, draw, thick, minimum width=0.64cm, minimum height=0.45cm] at (cv2_11) {};
\node[blue!10, right=0.05cm of cv2_11, fill, minimum width=0.64cm, minimum height=0.45cm] (cv2_12) {};
\node[blue, draw, thick, minimum width=0.64cm, minimum height=0.45cm] at (cv2_12) {};
\node[blue!10, right=0.05cm of cv2_12, fill, minimum width=0.64cm, minimum height=0.45cm] (cv2_13) {};
\node[blue, draw, thick, minimum width=0.64cm, minimum height=0.45cm] at (cv2_13) {};

\node[blue!10, below=0.05cm of cv2_10, fill, minimum width=0.64cm, minimum height=0.45cm] (cv2_20) {};
\node[blue, draw, thick, minimum width=0.64cm, minimum height=0.45cm] at (cv2_20) {};
\node[blue!10, right=0.05cm of cv2_20, fill, minimum width=0.64cm, minimum height=0.45cm] (cv2_21) {};
\node[blue, draw, thick, minimum width=0.64cm, minimum height=0.45cm] at (cv2_21) {};
\node[blue!10, right=0.05cm of cv2_21, fill, minimum width=0.64cm, minimum height=0.45cm] (cv2_22) {};
\node[blue, draw, thick, minimum width=0.64cm, minimum height=0.45cm] at (cv2_22) {};
\node[blue!10, right=0.05cm of cv2_22, fill, minimum width=0.64cm, minimum height=0.45cm] (cv2_23) {};
\node[blue, draw, thick, minimum width=0.64cm, minimum height=0.45cm] at (cv2_23) {};

\node[below right=0.25cm and -0.5cm of cv0_20]  {4D Correlation Volumes};


\node[inner sep=0pt, below right=0.45cm and 0.1cm of fcon_1_l4] (context_anchor) {};
\node[draw, minimum width=1cm, minimum height=1.5cm] (gruk) at (10, 0) {};
\node[left=10cm of gruk] (input_hidden) {Hidden State};
\draw[->, thick] (input_hidden.east) -- (gruk.west);

\node[inner sep=0pt, above=1cm of gruk, orange, draw, fill=orange!10, circle, minimum size=0.5cm] (lookupk) {\small $L'$ \normalsize};
\node[inner sep=0pt, right=0.1cm of lookupk] () {ALO};

\node[inner sep=0pt, below right=0.37cm of gruk, draw, circle] (flow_iter_k) {$+$};
\node[inner sep=0pt, left=6cm of flow_iter_k] (flow_k_minus_1) {$f$};
\node[inner sep=0pt, above=0.3cm of flow_iter_k] (delta_flow_k) {\tiny $\Delta f$ \normalsize};
\draw[->, thick] (flow_k_minus_1.east) -- (flow_iter_k.west);

\draw[->, thick] ([yshift=0.25cm] gruk.south east) .. controls  ([xshift=0.25cm, yshift=0.25cm] gruk.south east) and ([yshift=0.1cm] flow_iter_k.north) .. (flow_iter_k.north);
\draw[->, thick] (flow_iter_k.east) -- +(1, 0);

\draw[->, thick] (cv2_13.east) -| (lookupk.north);
\draw[->, thick] (lookupk.south) -- (gruk.north);

\node[draw=black, fill=white, thick, inner sep=4pt] (cat1) at (2, -4) {Concat};
\draw[->, thick] (input_hidden.east) -| (cat1.north);
\draw[->, dotted, thick] (context_anchor.east) -| (cat1.south);
\node[draw=green, right=0.5cm of cat1, fill=green!10, thick, minimum width=0.25cm, minimum height=1.5cm, label={[align=center]below:\small Conv. $1 \times 1$ \normalsize}] (conv) {};
\draw[->, thick] (cat1.east) -- (conv.west);

\node[right=1.75cm of conv, inner sep=0pt] (split) {};
\node[draw=black, above=of split, fill=white, thick, inner sep=4pt] (global_max_pool) {\tiny Global Max Pooling \normalsize};
\node[draw=black, below=of split, fill=white, thick, inner sep=4pt] (global_min_pool) {\tiny Global Min Pooling \normalsize};
\draw[->, thick] (conv.east) -| (global_max_pool.south);
\draw[->, thick] (conv.east) -| (global_min_pool.north);

\node[draw=black, right=of split, fill=white, thick, inner sep=4pt] (cat2) {Concat};
\draw[->, thick] (global_max_pool.east) -| (cat2.north);
\draw[->, thick] (global_min_pool.east) -| (cat2.south);

\node[draw=purple, above right=0.25 and 0.5cm of cat2, fill=purple!10, thick, minimum width=0.25cm, minimum height=1cm] (ff1) {};
\node[draw=purple, below right=0.25 and 0.5cm of cat2, fill=purple!10, thick, minimum width=0.25cm, minimum height=1cm, label={[align=center]below:\small Feed Forward Network \normalsize}] (ff2) {};

\node[right=0.125cm of cat2, inner sep=0pt] (split_again) {};

\draw[-, thick] (cat2.east) -- (split_again.east);
\draw[->, thick] (split_again.east) |- (ff1.west);
\draw[->, thick] (split_again.east) |- (ff2.west);

\node[draw=black, below=2.85cm of gruk, fill=white, thick, inner sep=4pt] (cat3) {Concat};
\node[inner sep=0, left=0.5cm of lookupk] (lookupk_left_anchor) {};
\draw[-, thick] (ff1.east) -| (lookupk_left_anchor.west);
\draw[-, thick] (ff2.east) -| (lookupk_left_anchor.west);
\draw[->, thick] (lookupk_left_anchor.west) -- (lookupk.west);

\draw[->, dotted, thick] (context_anchor.east) -| (cat3.south);
\draw[->, dotted, thick] (ff1.east) -- (cat3.north west);
\draw[->, dotted, thick] (ff2.east) -- (cat3.south west);
\draw[->, dotted, thick] (cat3.north) -- (gruk.south);
\end{tikzpicture}
}%
\caption{Scalar parameter calculation network for the ALO.}
\label{fig:alo_architecture}
\end{figure}
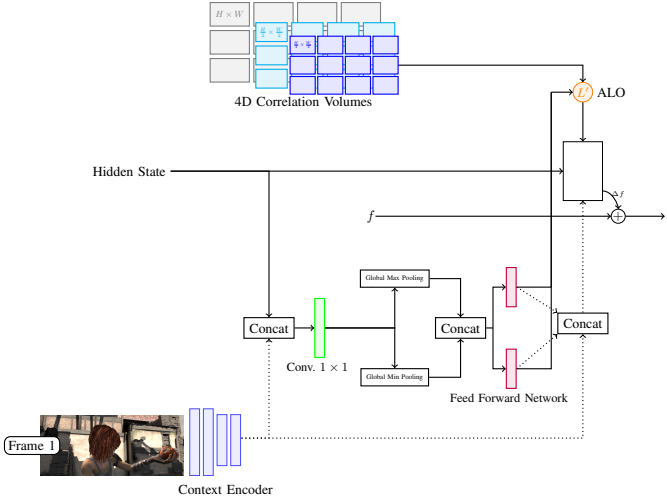

The aforementioned scalars are calculated via the application of a light-weight network module similar in structure to the BAM\cite{park2018bam} and CBAM\cite{woo2018cbam} architectures. The structure of this module is depicted in Fig.~\ref{fig:alo_architecture}. The network first concatenates the current hidden state with the context features, and enriches them with a $1 \times 1$ convolution. Then, it generates a global and compact descriptor of the algorithm's state by doing a Max and a Min pooling operation, and concatenating the resultant vectors. It then feeds this descriptor into two single layer, fully connected networks, each with a Sigmoid activation function. These networks then apply an appropriate affine transform on the result of the Sigmoid function, outputting the scalar for all levels of the correlation volume. The final scalars are then fed into the ALO, and also concatenated with the context features of the network. Concatenation with the context feature map is crucial, as the GRU will not know what type of ALO is being used otherwise, and cannot reap the benefits of having this new lookup operator.

Therefore, the light weight of this structure allows the network to calculate the scalar parameters in a memory and compute efficient manner, adapting the lookup operator based on the input image and the current stage of the regression algorithm.

\subsection{Attention-based Feature Localizer}
The vanilla RAFT architecture uses the outputs of the feature encoders to calculate the correlation volumes. Due to the convolutional structure of these encoders, the output features are inherently local, which may cause the network as whole not being able to differentiate between points in poorly textured regions of the frames. Consequently, the regression algorithm will have a hard time estimating a correct flow value for these regions of the frames, as it has to utilize the flow values outside of the regions to slowly correct the prediction of the region itself.

\begin{figure}
\centering
\resizebox{0.5\columnwidth}{!}{%
\begin{tikzpicture}
\draw[blue, pattern=checkerboard, pattern color=blue!10, rounded corners] (-3, -1) --
 (-2, 0.5) -- (-1, -0.25) -- (-1.5, 1) -- (-2.5, 1.1) -- (0.5, 2) -- (1, 0.25) -- (3, 1) -- (2.5, -0.5) -- (1, -0.25) -- (1.5, -1.25) -- (0.5, -1.5) -- (-1.5, -1.65) -- (-1.5, -0.3) -- cycle;
 
\node[inner sep=0pt, circle, draw, fill=black] at (0, 0) (p) {};
\node[above right=0 of p] {$p$};

\draw[-, red, very thick, decorate, decoration={brace, mirror, aspect=0.5}] (p.west) -- +(-1.07, 0) node[pos=0.5, above=0pt, black]{$x^-_p$};
\draw[-, red, very thick, decorate, decoration={brace, aspect=0.75}] (p.east) -- +(2.65, 0) node[pos=0.75, above=0pt, black]{$x^+_p$};
\draw[-, red, very thick, decorate, decoration={brace, mirror, aspect=0.5}] (p.north) -- +(0, 1.83) node[pos=0.5, right=0pt, black]{$y^-_p$};
\draw[-, red, very thick, decorate, decoration={brace, aspect=0.5}] (p.south) -- +(0, -1.52) node[pos=0.5, right=0pt, black]{$y^+_p$};

\draw[-, orange, ultra thick] (-1.35, 0) -- +(-1, 0);
\end{tikzpicture}
}%
\caption{Definition of the $x_p^{\pm}$ and $y_p^{\pm}$ values for a pixel $p$ in a poorly textured region. The orange points depict the pixels that may cause us to err in our estimation of $\Delta x_p$ and $\Delta y_p$.}
\label{fig:repetitive_coords}
\end{figure}
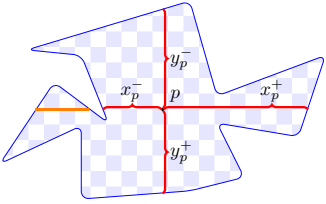

To remedy this, the aforementioned features are augmented with a set of ``global features" that encapsulate some data from across the entire frames, enabling the network to differentiate the poorly textures pixels by looking at these global features. To achieve this, for each pixel $p$ of each frame, the following four parameters are defined: $x_p^+$, $x_p^-$, $y_p^+$, and $y_p^-$, which represent how many pixels must we move from $p$ to the right (resp. left, up, and down) to exit the poorly textured region. An example of these values is depicted in Fig.~\ref{fig:repetitive_coords}.

The idea then is to not estimate the values $x_p^{\pm}$ and $y_p^{\pm}$ themselves, but rather estimate $\Delta x_p = x_p^+ - x_p^-$ and $\Delta y_p = y_p^+ - y_p^-$, which essentially represent the whereabouts of the point $p$ when compared to the ``middle-points" of the poorly textured region. Therefore, $\Delta x_p$ and $\Delta y_p$ create something similar to a coordinate system for each poorly textured shape. One can expect the values of $\Delta x_p$ and $\Delta y_p$ to not change by much for corresponding points, when poorly textured regions are translated only.

Suppose the values $\Delta x_p$ are calculated for the pixels in row $r$ of the first frame. To this end, the multi-head attention mechanism is applied on the features resulting from the feature encoders of this frame, which we shall denote $f_{r,i}$ for the $i$-th pixel of this row. Then, the keys and values for a query $q_i = f_{r, i}$ are defined as
\begin{equation}
    k_j = f_{r, j},
    v_j = pe(i - j),
\end{equation}
where $pe(\cdot)$ is the positional encoding function defined in\cite{vaswani2017attention} as $pe(\cdot)_{2k} = sin(\frac{\cdot}{1000^{k/d}})$, $pe(\cdot)_{2k + 1} = cos(\frac{\cdot}{1000^{k/d}})$. 

Since the keys $k_i$ and queries $q_i$ are defined in the same manner, in the context of a poorly textured region, this mechanism will take a weighted average of the relative positional encodings of the row, with the pixels in the region having a much larger weight that the pixels outside. This will result in an estimate of $\Delta x_p$ in the form of a positional encoding, which gives the network some idea as to what the actual value of $\Delta x_p$ is. The implementation uses four heads during this calculation, so that it can take into account more complex similarities between the feature vectors in the pooling procedure.

Notice that $v_j$ is a ``relative positional encoding" dependent on both $i$ and $j$. To calculate this relative positional encoding, first  the multi-head attention mechanism is applied using $v'_j = pe(j)$, which is the standard positional encoding, and then exploit the linearity of the attention mechanism along with well-known trigonometric identities to convert the input $j$ into $i - j$. It is worth noting that the resulting estimation is not without error. For instance, in Fig.~\ref{fig:repetitive_coords}, the points colored in orange will bias the estimate of $\Delta x_p$ towards the left.

This same procedure can be applied on all rows and columns of both frames to derive estimates for all values $\Delta x_p$ and $\Delta y_p$, all using the same module. The resulting estimates are then concatenated with the features of their corresponding frames, effectively augmenting the old features with a set of global features that span the entire row and column of each pixel. This resulting estimate lends itself well to how the correlation volume is calculated, as closer values of the input to $pe(\cdot)$ correspond to a larger dot-product, and thus more similarity.

It is worth noting that this mechanism is applied only on the rows and columns, enabling it to enjoy excellent speed and memory efficiency.

\begin{figure*}[!ht]
\centerline{\includegraphics[scale=.45]{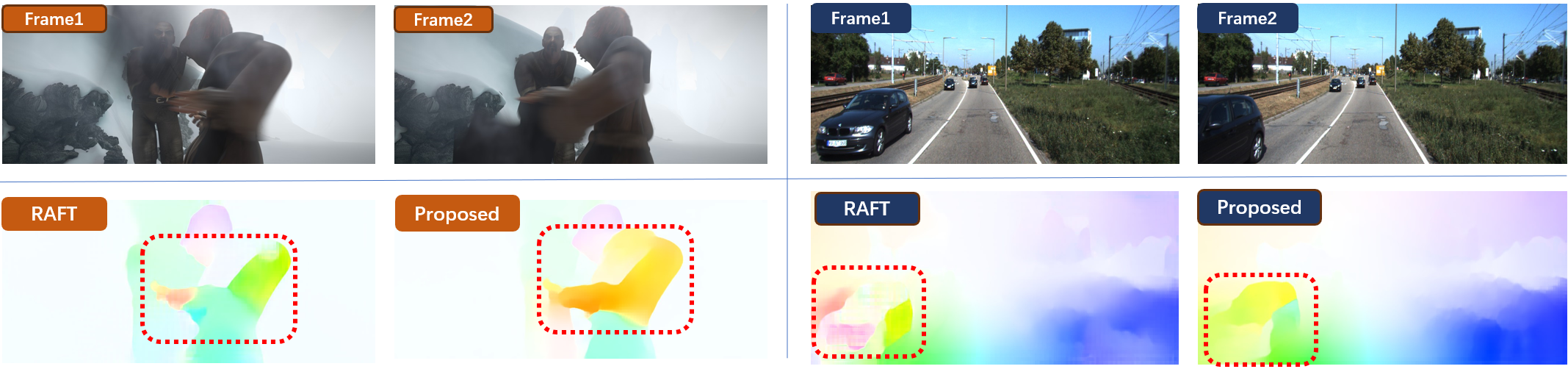}}
\caption{Qualitative comparison between the proposed method and RAFT. Frames with orange and blue labels are from Sintel and KITTI datasets, respectively.}
\label{fig:2_Visualization}
\end{figure*}

\section{EXPERIMENTS}
\subsection{Datasets and Training Schedule}
Based on prior works, the model is initially pretrained on the FlyingChairs dataset\cite{dosovitskiy2015flownet} for 100k iterations using a batch size of 10. Following that, it undergoes an additional pretraining phase on the FlyingThings dataset\cite{mayer2016large} for 200k iterations with a batch size of 6. In this work, no fine-tuning is performed. The evaluation is conducted on both the train split of the KITTI-15\cite{geiger2013vision} dataset and the train split of the MPI Sintel\cite{butler2012naturalistic} dataset, considering both the Clean and Final passes of the data.

For the ablation results, fast configuration was employed for training and evaluation purposes. In this configuration, the model underwent training for 25k steps using the train set of the MPI Sintel dataset. Following the training phase, the model was evaluated on the train set of the KITTI-15 dataset.

The implementations of all the networks were done using the PyTorch framework. Training and inference processes were conducted on a single NVIDIA GeForce RTX 3090 GPU. The training process involves the utilization of the AdamW optimizer with a one-cycle learning rate scheduler. Additionally, the gradients are clipped to the range of [-1, 1]. The hyperparameter $\gamma$ in the equation \ref{Loss Function} is set to 0.8, similar to the approach used in\cite{teed2020raft}.


\subsection{Evaluation Metrics}
The main evaluation metric for the MPI Sintel dataset is the end-point error (EPE), which represents the average error of the flow estimation at each pixel, measured in terms of the number of pixels. For the KITTI-15 dataset, the evaluation metrics used are F1-EPE and F1-All. F1-EPE is the same as the EPE metric described for the Sintel dataset. On the other hand, F1-All measures the percentage of outliers, specifically pixels whose end-point error exceeds either 3 pixels or 5\% of the ground truth flow magnitude. This metric is averaged over all pixels in the dataset. To validate the efficiency of the proposed method, additional metrics such as the number of parameters in the network, runtime, and memory usage are reported.

\subsection{Experimental Results}
Extensive experiments were conducted to validate the effectiveness of the proposed method by comparing it with several existing methods, including RAFT used as the baseline. The results of the proposed method in comparison to these existing methods are presented in Table \ref{tab:results}. The table demonstrates that the proposed method achieves great results in terms of EPE on the challenging Sintel dataset, particularly on the Final pass. Moreover, on the KITTI dataset and the F1-All evaluation metric, the proposed method achieves the third-best performance. It is worth noting that GMFlowNet, being a transformer-based method, requires a larger amount of training data and has a higher number of parameters compared to our method. In light of this, the proposed method still achieves remarkable results overall. Specifically, compared to RAFT, it demonstrates more than a 10\% improvement on the Sintel dataset and nearly a 6\% improvement on the KITTI dataset, which means the proposed method plays a significant role in improving large errors.

Additionally, it is important to mention that CRAFT and GMA are two methods that exhibit results comparable to the proposed method. These methods have pursued independent approaches in their research and can even be combined with the proposed method to further enhance performance.

Table \ref{tab:runtime} presents a runtime comparison between the proposed method and RAFT, considering two different numbers of steps. The table illustrates that the proposed method improves upon RAFT while introducing only a 33\% runtime overhead. This indicates a balanced trade-off between accuracy and efficiency, as the proposed method achieves better results while maintaining reasonable runtime performance.

Table \ref{tab:memmory} provides a comparison of RAFT, the proposed method, and GMFlowNet in terms of the number of parameters and memory usage. As mentioned earlier, the proposed method utilizes significantly less GPU memory compared to GMFlowNet, making it feasible to run on the 24GB RAM of the Nvidia GeForce 3090 GPU. Additionally, the table reveals that the proposed method introduces only a slight increase in the number of parameters and memory usage (13\%) compared to RAFT, despite its significant improvement in flow estimation accuracy. This highlights the efficiency of the proposed method in achieving impressive results with relatively fewer parameters and memory requirements.

\begin{table}[!ht]
  \centering
  \caption{Comparison of the proposed method with existing techniques on the Sintel and KITTI datasets. \textcolor{green}{Green}, \textcolor{blue}{blue}, and \textcolor{red}{red} colors denote the first, second, and third-best results.}
    \begin{tabular}{|c|c c|c c|}
    \hline
    {\multirow{2}{*}{\makecell[c]{Method}}} & \multicolumn{2}{c|}{MPI Sintel (Train)} & \multicolumn{2}{c|}{KITTI-15 (Train)}  \\
    \cline {2-5}
    & EPE (Clean)  & EPE (Final)  & F1-EPE  & F1-All  \\
    \hline

    
    PWC-Net\cite{sun2018pwc} & \multicolumn{1}{c|}{2.55} & 3.93 & \multicolumn{1}{c|}{10.35} & \multicolumn{1}{c|}{33.7}\\

    LiteFlowNet\cite{hui2018liteflownet} & \multicolumn{1}{c|}{2.48} & 4.04 & \multicolumn{1}{c|}{10.39} & \multicolumn{1}{c|}{28.5}\\
    
    LiteFlowNet2\cite{hui2020lightweight} & \multicolumn{1}{c|}{2.24} & 3.78 & \multicolumn{1}{c|}{8.97}  & \multicolumn{1}{c|}{25.9}\\

    VCN\cite{yang2019volumetric} & \multicolumn{1}{c|}{2.21} & 3.68 & \multicolumn{1}{c|}{8.36}  & \multicolumn{1}{c|}{25.1}\\

   MaskFlowNet\cite{zhao2020maskflownet}  & \multicolumn{1}{c|}{2.25} & 3.61  & \multicolumn{1}{c|}{-} & \multicolumn{1}{c|}{23.1}\\
    
   FlowNet2\cite{ilg2017flownet}  & \multicolumn{1}{c|}{2.02} & 3.54  & \multicolumn{1}{c|}{10.08}  & \multicolumn{1}{c|}{30.0}\\

   DICL\cite{wang2020displacement}  & \multicolumn{1}{c|}{1.94} & 3.77  & \multicolumn{1}{c|}{8.70}  & \multicolumn{1}{c|}{23.60}\\

   Flow1D\cite{xu2021high}  & \multicolumn{1}{c|}{1.98} & 3.27  & \multicolumn{1}{c|}{5.59}  & \multicolumn{1}{c|}{22.95}\\

   RAFT\cite{teed2020raft}  & \multicolumn{1}{c|}{1.43} & 2.71  & \multicolumn{1}{c|}{5.02}  & \multicolumn{1}{c|}{17.46}\\

   FM-RAFT\cite{jiang2021learning2}  & \multicolumn{1}{c|}{1.29} & 2.95  & \multicolumn{1}{c|}{6.98}  & \multicolumn{1}{c|}{19.3}\\

   CRAFT\cite{sui2022craft}  & \multicolumn{1}{c|}{\textcolor{blue}{1.27}} & 2.79  & \multicolumn{1}{c|}{4.88}  & \multicolumn{1}{c|}{17.50}\\

   GMA\cite{jiang2021learning}  & \multicolumn{1}{c|}{1.30} & 2.74  & \multicolumn{1}{c|}{4.69}  & \multicolumn{1}{c|}{17.10}\\

   AGFlow\cite{luo2022learning}  & \multicolumn{1}{c|}{1.31} & 2.69  & \multicolumn{1}{c|}{4.82}  & \multicolumn{1}{c|}{17.0}\\

   KPA-Flow\cite{luo2022learning2}  & \multicolumn{1}{c|}{\textcolor{red}{1.28}} & 2.68  & \multicolumn{1}{c|}{\textcolor{blue}{4.46}}  & \multicolumn{1}{c|}{\textcolor{blue}{15.90}}\\

   S-Flow\cite{zhang2021separable} & \multicolumn{1}{c|}{1.30} & \textcolor{green}{2.59} & \multicolumn{1}{c|}{\textcolor{red}{4.60}}  & \multicolumn{1}{c|}{\textcolor{blue}{15.90}}\\

   GMFlowNet\cite{zhao2022global}  & \multicolumn{1}{c|}{\textcolor{green}{1.14}} & 2.71  & \multicolumn{1}{c|}{\textcolor{green}{4.24}}  & \multicolumn{1}{c|}{\textcolor{green}{15.40}}\\

   EMD-S\cite{deng2023explicit} & \multicolumn{1}{c|}{1.31} &\textcolor{red}{ 2.67} & \multicolumn{1}{c|}{5.0}  & \multicolumn{1}{c|}{17.0}\\

   \textbf{Ef-RAFT (ours)}  & \multicolumn{1}{c|}{\textcolor{blue}{1.27}} & \textcolor{blue}{2.60}  & \multicolumn{1}{c|}{4.83}  & \multicolumn{1}{c|}{\textcolor{red}{16.45}}\\


    \hline
    \end{tabular}%
  \label{tab:results}%
\end{table}%
\begin{table}[ht]
  \centering
  \caption{Runtime comparison between the proposed method and RAFT.}
  \label{tab:runtime}
  \renewcommand{\arraystretch}{1.2}
  \begin{tabular}{|c|c|c|c|c|}
    \hline
    \textbf{Method} & \textbf{\#Steps} & \textbf{Runtime (s)} & \textbf{EPE (Clean)} & \textbf{EPE (Final)} \\
    \hline
    RAFT & 32 & 87.81 & 1.46 & 2.69 \\
    \hline
    Proposed & 32 & 117.64 & 1.29 & 2.62 \\
    \hline
    RAFT & 22 & 60.39 & 1.48 & 2.69 \\
    \hline
    Proposed & 22 & 85.91 & 1.28 & 2.60 \\
    \hline
  \end{tabular}
\end{table}
\begin{table}[ht]
  \centering
  \caption{Number of parameters and memory usage comparison.}
  \label{tab:memmory}
  \renewcommand{\arraystretch}{1.2}
  \begin{tabular}{|c|c|c|c|}
    \hline
    \textbf{Method} & \textbf{\#Steps} & \textbf{\#Parameters} & \textbf{Memmory Used} \\
    \hline
    RAFT & 12 & 5.3 M & 11988 (MiB) \\
    \hline
    GMFlowNet & 12 & 9.3 M & $\ge$ 24000 (MiB)  \\
    \hline
    RAFT & 12 & 5.7 M & 13520 (MiB) \\
    \hline

  \end{tabular}
\end{table}

\subsection{Ablation}
The effectiveness of the proposed components was further validated through an ablation study. Quick training was conducted, where the lower steps method was trained on the Sintel dataset and evaluated on the KITTI dataset. The results in Table \ref{tab:ablation} show that both ALO and AFL components contribute to the improvement over RAFT. While the AFL alone achieves better results on the KITTI dataset, the combined inclusion of both components is believed to offer greater learning ability. Overall, the study demonstrates that the synergistic effect of incorporating both components leads to enhanced performance compared to individual components.

\begin{table}[!ht]
  \centering
  \caption{Ablation for Ef-RAFT to validate the effectiveness of proposed ALO and AFL components.}
    \begin{tabular}{|c|c c|c c|}
    \hline
    {\multirow{2}{*}{\makecell[c]{Method}}} & \multicolumn{2}{c|}{MPI Sintel (Train)} & \multicolumn{2}{c|}{KITTI-15 (Train)}  \\
    \cline {2-5}
    & EPE (Clean)  & EPE (Final)  & F1-EPE  & F1-All  \\
    \hline
    RAFT\cite{teed2020raft} & \multicolumn{1}{c|}{1.73} & 2.40 & \multicolumn{1}{c|}{8.14} & \multicolumn{1}{c|}{22.60}\\
    
    w/o AFL & \multicolumn{1}{c|}{1.57} & 2.19 & \multicolumn{1}{c|}{6.40}  & \multicolumn{1}{c|}{18.81}\\


   w/o ALO  & \multicolumn{1}{c|}{1.81} & 2.41  & \multicolumn{1}{c|}{\textbf{3.15}} & \multicolumn{1}{c|}{\textbf{11.36}}\\
    
   Proposed  & \multicolumn{1}{c|}{\textbf{1.57}} & \textbf{2.13}  & \multicolumn{1}{c|}{6.02}  & \multicolumn{1}{c|}{18.75}\\

    \hline
    \end{tabular}%
  \label{tab:ablation}%
\end{table}%

\subsection{Qualitative Assessment}
Qualitative assessment was conducted to further validate the proposed method. Figure \ref{fig:2_Visualization} demonstrates that the proposed method is capable of estimating flows that are significantly better than RAFT. The visual comparison clearly showcases the improved accuracy and quality of the flow estimations achieved by the proposed method.

\section{CONCLUSION AND FUTURE WORK}
In this paper, Ef-RAFT was presented, a re-imagining of the renowned RAFT network. The novel ideas of Amorphous Lookup Operator and Attention-based Feature Localizer were explored, which enabled Ef-RAFT to improve upon its predecessor while also keeping the computational complexity and memory footprint low. Then experiments were conducted to showcase the improved accuracy and efficiency of Ef-RAFT, as well as demonstrating the necessity of the structures used ith an ablation study. It is believed that Ef-RAFT can be extended in several directions: 1) The Attention-Based Feature Localizer is not robust to rotations in the poorly textured regions. Applying the same mechanism on rotated lines in the images, instead of the rows and columns alone, may be a promising first step forward. 2) Ef-RAFT's ideas are orthogonal to what is used in many other papers, e.g. CRAFT and GMA. Combining these methods in an efficient and practical manner may also open doors to better accuracies with a light-weight network.

\bibliographystyle{IEEEtran}
\bibliography{main}

\end{document}